\pdfoutput=1
\documentclass{article}
\usepackage{ijcai22}
\usepackage{times}
\usepackage{soul}
\usepackage[small]{caption}
\usepackage[bookmarks=false]{hyperref}
\hypersetup{
    final, hidelinks,
    pdfauthor={Martin Isaksson, Edvin Listo Zec, Rickard Cöster, Daniel Gillblad and Sarunas Girdzijauskas},
    pdftitle={Adaptive Expert Models for Personalization in Federated Learning}
}

\usepackage{graphicx}

\usepackage[export]{adjustbox}
\usepackage{flushend}
\usepackage{pgfplots}

\DeclareUnicodeCharacter{2212}{−}
\usepgfplotslibrary{groupplots,dateplot}
\pgfplotsset{compat=newest}

\usepackage{url}
\usepackage{amsmath}
\usepackage{amssymb}
\usepackage{bm}
\usepackage{mathtools}
\usepackage[operators]{cryptocode}
\usepackage[misc,geometry]{ifsym}
\usepackage{tikz, adjustbox, tikz-qtree}
\usepackage{balance}
\usepackage{algorithm}
\usepackage[noend]{algpseudocode}         
\usepackage{microtype}
\usepackage{subfig}

\usepackage[capitalise]{cleveref}  

\Crefname{algorithm}{Alg.}{Algs.}%
\crefname{algorithm}{Algorithm}{Algorithms}%
\crefname{table}{Table}{Tables}
\crefname{figure}{Figure}{Figures}

\usepackage{booktabs}              
\usepackage[round-mode=places,detect-weight=true,detect-inline-weight=math]{siunitx}  

\usepackage{multirow}
\usepackage{array}
\usepackage{arydshln}

\usepackage[inline]{enumitem}
\setlist{noitemsep}

\DeclareSIUnit\px{px}
\DeclareSIUnit\samples{data~samples}

\microtypecontext{spacing=nonfrench}
\microtypesetup{
    activate={true, nocompatibility},
    babel=false,
    final,
    tracking=false,
    kerning=true,
    spacing=true,
    factor=1100,
    shrink=30}
\SetTracking{encoding={*}, shape=sc}{40}

\usepackage[graph, grays]{ericolors}
\usepackage{todonotes}

\usetikzlibrary{patterns,shapes.arrows, arrows.meta, fit, calc, intersections}
\usetikzlibrary{shapes, decorations, arrows}
\usetikzlibrary{3d,fit,backgrounds, decorations.text}
\usetikzlibrary{positioning, shapes.symbols}
\usetikzlibrary{decorations.pathreplacing, calligraphy}

\tikzset{>=latex}

\pgfplotsset{compat=newest,
    width=6cm,
    height=2.5cm,
    scale only axis=true,
    max space between ticks=25pt,
    try min ticks=5,
    every axis/.style={
        axis y line=left,
        axis x line=bottom,
        axis line style={thick,->,>=latex, shorten >=-.4cm},
        mark size=1,
        label style={font=\large},
        tick label style={font=\large}
    },
    every axis plot/.append style={thick},
    tick style={black, thick},
    colorbar style={at={(1.25, 1)}}
}
\tikzset{
    semithick/.style={line width=0.8pt},
}

\newcommand{\suma}{\Large$+$}
\newcommand{\R}{\mathbb{R}}
\newcommand{\N}{\mathbb{N}}

\def\gatesep{.3}

\usepackage[acronym,section=subsection,symbols]{glossaries-extra}
\setabbreviationstyle[acronym]{long-short}
\glsdisablehyper

\let\originalmiddle=\middle%
\def\middle#1{\mathrel{}\originalmiddle#1\mathrel{}}

\DeclareMathOperator{\Prob}{\mathcal{P}}
\DeclareMathOperator*{\argminB}{argmin}
\newcommand{\flconstfnt}[1]{\ensuremath{\textsc{#1}}}

\newcommand{\cifar}{\mbox{\gls{CIFAR}-10}}

\newcommand{\squeezeup}{\vspace{-.5\baselineskip plus 2pt minus 1pt}}
\newacronym{3GPP}{3GPP}{Third Generation Partnership Project}
\newacronym{ANN}{ANN}{Artificial Neural Network}
\newacronym{API}{API}{Application Programming Interface}
\newacronym{CA}{CA}{Certificate Authority}
\newacronym{CDF}{CDF}{Cumulative Distribution Function}
\newacronym{CIFAR}{CIFAR}{Canadian Institute For Advanced Research}
\newacronym{CNN}{CNN}{Convolutional Neural Network}
\newacronym{DEF}{DEF}{Data Expansion Factor}
\newacronym{DoS}{DoS}{Denial-of-service}
\newacronym{ECIES}{ECIES}{Elliptic Curve Integrated Encryption Scheme}
\newacronym{FedAvg}{\textsc{FedAvg}}{Federated Averaging}
\newacronym{FEMNIST}{FEMNIST}{Federated Extended MNIST}
\newacronym{FL}{FL}{Federated Learning}
\newacronym{FQDN}{FQDN}{Fully Qualified Domain Name}
\newacronym{GPU}{GPU}{Graphics Processing Unit}
\newacronym{IID}{IID}{Independent and Identically Distributed}
\newacronym{IFCA}{IFCA}{Iterative Federated Clustering Algorithm}
\newacronym{IoT}{IoT}{Internet of Things}
\newacronym{IP}{IP}{Internet Protocol}
\newacronym{KDF}{KDF}{Key Derivation Function}
\newacronym{KPI}{KPI}{Key Performance Indicator}
\newacronym{KTH}{KTH}{KTH Royal Institute of Technology}
\newacronym{MAB}{MAB}{Multi-Armed Bandit}
\newacronym{MAC}{MAC}{Message Authentication Code}
\newacronym{ML}{ML}{Machine Learning}
\newacronym{MoE}{MoE}{Mixture of Experts}
\newacronym{MPC}{MPC}{Multi-Party Computation}
\newacronym{PKI}{PKI}{Public Key Infrastructure}
\newacronym{PLMN}{PLMN}{Public Land Mobile Network}
\newacronym{PRF}{PRF}{Pseudo Random Function}
\newacronym{PRG}{PRG}{Pseudo Random Generator}
\newacronym{PS}{PS}{Public Safety}
\newacronym{PRNG}{PRNG}{Pseudo Random Number Generator}
\newacronym{QoE}{QoE}{Quality of Experience}
\newacronym{RAN}{RAN}{Radio Access Network}
\newacronym{RSA}{RSA}{Rivest-Shamir-Adleman}
\newacronym{RSRP}{RSRP}{Reference Signal Received Power}
\newacronym{SNIC}{SNIC}{Swedish National Infrastructure for Computing}
\newacronym{SAR}{SAR}{Search and Rescue}
\newacronym{SGD}{SGD}{Stochastic Gradient Descent}
\newacronym{TCP}{TCP}{Transmission Control Protocol}
\newacronym{TLS}{TLS}{Transport Layer Security}
\newacronym{UDP}{UDP}{User Datagram Protocol}
\newacronym{UE}{UE}{User Equipment}
\newacronym{UPF}{UPF}{User Plane Function}
\newacronym{UAV}{UAV}{Unmanned Aerial Vehicle}
\newacronym{USV}{USV}{Unmanned Surface Vehicle}
\newacronym{UCB}{UCB}{Upper Confidence Bound}
\newacronym{UUID}{UUID}{Universally Unique Identifier}
\newacronym{WARA-PS}{WARA-PS}{WASP Autonomous Research Arenas --- Public Safety}
\newacronym{WARA}{WARA}{WASP Autonomous Research Arenas}
\newacronym{WASP}{WASP}{Wallenberg AI, Autonomous Systems and Software Program}

\newacronym{AF}{AF}{Application Function}
\newacronym{NF}{\text{NF}}{Network Function}
\newacronym{NRF}{NRF}{NF Repository Function}
\newacronym{NWDA}{NWDA}{Network Data Analytics}
\newacronym{NWDAF}{NWDAF}{Network Data Analytics Function}
\newacronym{AMF}{AMF}{Access and Mobility Management Function}
\newacronym{AUSF}{AUSF}{AUthentication Server Function}
\newacronym{OAM}{OAM}{Operation, Administration, and Maintenance}
\newacronym{SBA}{SBA}{Service Based Architecture}

\newglossaryentry{eta}{name={\ensuremath{\eta}}, sort={eta}, description={Learning rate}, type={symbols}}
\newglossaryentry{B}{name={\ensuremath{B}}, sort={B}, description={Batch size}, type={symbols}}
\newglossaryentry{C}{name={\ensuremath{C}}, sort={C}, description={Fraction of clients selected}, type={symbols}}
\newglossaryentry{E}{name={\ensuremath{E}}, sort={E}, description={local epochs}, type={symbols}}

\newglossaryentry{eps}{name={\ensuremath{\varepsilon}}, sort={eps}, description={\(\varepsilon\)-greedy parameter}, type={symbols}}

\newglossaryentry{t}{name={\ensuremath{t}},sort={t},description={time in communication rounds},type={symbols}}
\newglossaryentry{flf}{name={\ensuremath{\mathcal{L}}},sort={L},description={Estimated total loss},type={symbols}}
\newglossaryentry{fl}{name={\ensuremath{f_l}},sort={fl},description={Local model},type={symbols}}
\newglossaryentry{flk}{name={\ensuremath{f^k_l}},sort={flk},description={Local model for client $k$},type={symbols}}
\newglossaryentry{flkp}{name={\ensuremath{f^{k'}_l}},sort={flkp},description={Local model for client $k'$},type={symbols}}
\newglossaryentry{fgj}{name={\ensuremath{f^j_g}},sort={fgj},description={Cluster model with index $j$},type={symbols}}
\newglossaryentry{fg}{name={\ensuremath{f_g}},sort={fgj},description={Global model},type={symbols}}
\newglossaryentry{Dn}{name={\ensuremath{D_n}},sort={dn},description={Dataset of length $n$},type={symbols}}
\newglossaryentry{p}{name={\ensuremath{p}}, sort={p}, description={majority class fraction}, type={symbols}}
\newglossaryentry{P}{name={\ensuremath{\Prob}}, sort={P}, description={probability}, type={symbols}}
\newglossaryentry{w}{name={\ensuremath{\bm{w}}}, sort={w}, description={model parameters}, type={symbols}}
\newglossaryentry{wk}{name={\ensuremath{\bm{w}_l}}, sort={wl}, description={Local model parameters}, type={symbols}}
\newglossaryentry{whk}{name={\ensuremath{\bm{w}^k_h}}, sort={whk}, description={Gating model parameters for client $k$}, type={symbols}}
\newglossaryentry{wlk}{name={\ensuremath{\bm{w}^k_l}}, sort={wlk}, description={Local model parameters for client $k$}, type={symbols}}
\newglossaryentry{wlt}{name={\ensuremath{\bm{w}_l(t)}}, sort={wlt}, description={Local model parameters at time $t$}, type={symbols}}
\newglossaryentry{wkt}{name={\ensuremath{\bm{w}^k(t)}}, sort={wkt}, description={Global model parameters for global model at time $t$}, type={symbols}}

\newglossaryentry{wkt1}{name={\ensuremath{\bm{w}^k(t+1)}}, sort={wlt}, description={Global model parameters for global model at time $t+1$}, type={symbols}}
\newglossaryentry{wg}{name={\ensuremath{\bm{w}_{g}}}, sort={wg}, description={Global model parameters}, type={symbols}}
\newglossaryentry{wgt}{name={\ensuremath{\bm{w}_{g}(t)}}, sort={wgt}, description={Global model parameters at time $t$}, type={symbols}}
\newglossaryentry{wg0}{name={\ensuremath{\bm{w}_{g}(0)}}, sort={wgt}, description={Global model parameters at time 0}, type={symbols}}
\newglossaryentry{wgt1}{name={\ensuremath{\bm{w}_{g}(t+1)}}, sort={wgt}, description={Global model parameters at time $t+1$}, type={symbols}}

\newglossaryentry{wgj}{name={\ensuremath{\bm{w}^j_g}}, sort={wgj}, description={Cluster model $j$ parameters}, type={symbols}}
\newglossaryentry{wgjt}{name={\ensuremath{\bm{w}^j_g(t)}}, sort={wgjt}, description={Cluster model $j$ parameters at time $t$}, type={symbols}}
\newglossaryentry{wgjht}{name={\ensuremath{\bm{w}^{\hat{j}}_g(t)}}, sort={wgjt}, description={Cluster model $\hat{j}$ parameters at time $t$}, type={symbols}}

\newglossaryentry{wgjt0}{name={\ensuremath{\bm{w}^j_g(0)}}, sort={wgjt0}, description={Cluster model $j$ parameters at time $0$}, type={symbols}}
\newglossaryentry{wgjt1}{name={\ensuremath{\bm{w}^j_g(t+1)}}, sort={wgjt1}, description={Cluster model $j$ parameters at time $t+1$}, type={symbols}}

\newglossaryentry{gl}{name={\ensuremath{g_l}},sort={gl},description={Gate model weight for local model},type={symbols}}
\newglossaryentry{glk}{name={\ensuremath{g_l^k}},sort={glk},description={Gate model weight for local model for client $k$},type={symbols}}
\newglossaryentry{g}{name={\ensuremath{g}},sort={g},description={Gate model weight},type={symbols}}
\newglossaryentry{gjk}{name={\ensuremath{g_j^k}},sort={gjk},description={Gate model weight for cluster model $j$ and client $k$},type={symbols}}

\newglossaryentry{h}{name={\ensuremath{f_h}},sort={h},description={Gate model function},type={symbols}}
\newglossaryentry{hk}{name={\ensuremath{f_h^k}},sort={hk},description={Gate model for client $k$},type={symbols}}
\newglossaryentry{hkp}{name={\ensuremath{f_h^{k'}}},sort={hk},description={Gate model for client $k$},type={symbols}}

\newglossaryentry{J}{name={\ensuremath{J}}, sort={J}, description={Number of cluster models}, type={symbols}}

\newglossaryentry{j}{name={\ensuremath{j}}, sort={j}, description={Cluster model index}, type={symbols}}
\newglossaryentry{jhat}{name={\ensuremath{\hat{j}}}, sort={jh}, description={Cluster model identity estimate}, type={symbols}}
\newglossaryentry{k}{name={\ensuremath{k}}, sort={k}, description={Client index}, type={symbols}}
\newglossaryentry{K}{name={\ensuremath{K}}, sort={K}, description={Number of clients}, type={symbols}}
\newglossaryentry{kp}{name={\ensuremath{k'}}, sort={kp}, description={Client index, primed}, type={symbols}}

\newglossaryentry{l}{name={\ensuremath{l}},sort={l},description={Loss function},type={symbols}}

\newglossaryentry{Ks}{name={\ensuremath{K_s}}, sort={ks}, description={Number of selected clients}, type={symbols}}
\newglossaryentry{nk}{name={\ensuremath{n_k}}, sort={nk}, description={Number of data samples for client $k$}, type={symbols}}
\newglossaryentry{nj}{name={\ensuremath{n_j}}, sort={nj}, description={Total number of data samples for cluster $j$ in one iteration.}, type={symbols}}

\newglossaryentry{n}{name={\ensuremath{n}}, sort={n}, description={Total number of data samples}, type={symbols}}

\newglossaryentry{x}{name={\ensuremath{\bm{x}}}, sort={x}, description={Features of a data sample}, type={symbols}}
\newglossaryentry{y}{name={\ensuremath{y}}, sort={y}, description={Target of a data sample}, type={symbols}}

\newglossaryentry{i}{name={\ensuremath{i}}, sort={i}, description={Data sample index}, type={symbols}}
\newglossaryentry{xi}{name={\ensuremath{\bm{x}_i}}, sort={xi}, description={Features of data sample $i$}, type={symbols}}
\newglossaryentry{yi}{name={\ensuremath{y_i}}, sort={yi}, description={Target of data sample $i$}, type={symbols}}
\newglossaryentry{yhatl}{name={\ensuremath{\hat{y}_l}}, sort={yhatl}, description={Estimated target (local)}, type={symbols}}
\newglossaryentry{yhath}{name={\ensuremath{\hat{y}_h}}, sort={yhath}, description={Estimated target (gating)}, type={symbols}}
\newglossaryentry{yhatg}{name={\ensuremath{\hat{y}_g}}, sort={yhatg}, description={Estimated target (gating)}, type={symbols}}
\newglossaryentry{yhatj}{name={\ensuremath{\hat{y}_j}}, sort={yhatj}, description={Estimated target (cluster)}, type={symbols}}

\newglossaryentry{Pk}{name={\ensuremath{P^k}}, sort={Pk}, description={Partition of dataset accessible to client $k$}, type={symbols}}

\newglossaryentry{S}{name={\ensuremath{\left\{1,2,\ldots,K\right\}}}, sort={S}, description={Population of clients}, type={symbols}}
\newglossaryentry{St}{name={\ensuremath{S_t}}, sort={St}, description={Selected set of clients at time $t$}, type={symbols}}
\newglossaryentry{Jset}{name={\ensuremath{\left\{1,2,\ldots,J\right\}}}, sort={J}, description={$[J] = \left\{j \in \N^+ \,\middle\vert\, j \leq J\right\}$}, type={symbols}}
\newglossaryentry{Jremset}{name={\ensuremath{\mathcal{J}}}, sort={J}, description={Set of cluster models}, type={symbols}}

\newglossaryentry{Eset}{name={\ensuremath{\left\{1,2,\ldots,E\right\}}}, sort={J}, description={$[E] = \left\{e \in \N^+ \,\middle\vert\, e \leq E\right\}$}, type={symbols}}

\title{Adaptive Expert Models for Personalization in Federated Learning}

\author{
Martin~Isaksson$^{1,2}$\and
Edvin~Listo~Zec$^{3,2}$\and
Rickard~Cöster$^{1}$\and
Daniel~Gillblad$^{4,5}$\And
\v{S}ar\={u}nas~Girdzijauskas$^{3,2}$\\
\affiliations
$^1$Ericsson AB, Stockholm, Sweden\\
$^2$KTH Royal Institute of Technology, Stockholm, Sweden\\
$^3$RISE Research Institutes of Sweden, Stockholm, Sweden\\
$^4$AI Sweden, Stockholm, Sweden\\
$^5$Chalmers AI Research Center, Chalmers University of Technology, Göteborg, Sweden%
}
\begin{document}
\maketitle
\global\csname @topnum\endcsname 0
\global\csname @botnum\endcsname 0

\begin{abstract}%
\gls{FL} is a promising framework for distributed learning when data is private
and sensitive. However, the state-of-the-art solutions in this framework are
not optimal when data is heterogeneous and non-\glsxtrshort{IID}\@. We propose a practical and
robust approach to personalization in \gls{FL} that adjusts to
heterogeneous and non-\glsxtrshort{IID} data by balancing exploration and
exploitation of several global models. To achieve our aim of personalization,
we use a \gls{MoE} that learns to group clients that are similar to each
other, while using the global models more efficiently.
We show that our approach achieves an accuracy
up to \SI{29.78}{\percent} better than the state-of-the-art
and up to \SI{4.38}{\percent} better compared to
a local model in a pathological non-\glsxtrshort{IID} setting, even though we tune our
approach in the \glsxtrshort{IID} setting.
\end{abstract}%

\glsresetall%

\section{Introduction}
\begin{figure}[t]
\centering
\includegraphics[width=.9\linewidth]{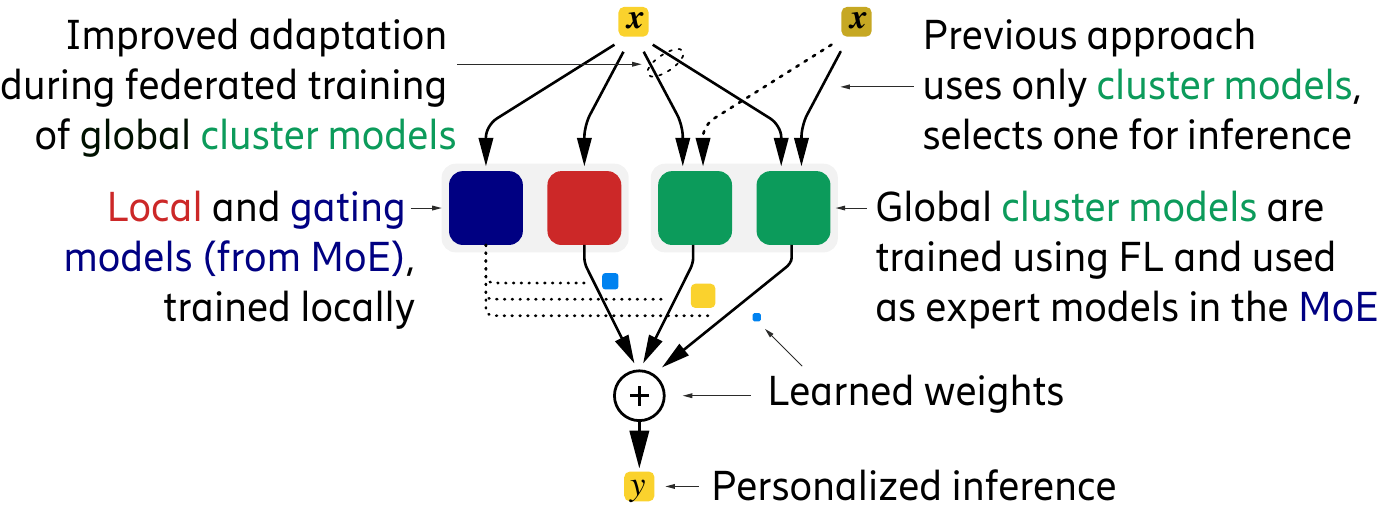}%
\caption{Our approach adjusts to non-\glsxtrfull{IID} data distributions by adaptively
training a \glsxtrfull{MoE} for clients that share similar data distributions.}\label{fig:overview}%
\end{figure}%
In many real-world scenarios, data is distributed over organizations or devices
and is difficult to centralize. Due to legal reasons, data might have to remain
and be processed where it is generated, and in many cases may not be allowed to 
be transferred~\cite{gdpr16}.
Furthermore, due to communication limitations it can be
practically impossible to send data to a central point of processing.
In many applications of \gls{ML} these challenges are becoming
increasingly important to address. For example, sensors, cars, radio
base stations and mobile devices are capable of generating
more relevant training data than can be practically communicated to the
cloud~\cite{ericssonaiwhitepaper}, and datasets in healthcare and industry
cannot legally be moved between hospitals or countries of origin.

\glsunset{IID}\gls{FL}~\cite{DBLP:conf/aistats/McMahanMRHA17,DBLP:conf/mlsys/BonawitzEGHIIKK19}
shows promise to leveraging data that cannot easily be
centralized. It has the potential to utilize
compute and storage resources of clients to scale towards large,
decentralized datasets while enhancing privacy. However, current
approaches fall short when data is heterogeneous as well as
non-Independent and Identically Distributed 
(non-\gls{IID}), where stark differences between clients
and groups of clients can be found. Therefore, personalization of collectively
learned models will in practice often
be critical to adapt to differences between regions, organizations and
individuals to achieve the required
performance~\cite{DBLP:journals/corr/abs-1912-04977,DBLP:conf/nips/GhoshCYR20}.
This is the problem we address in this paper.

Our approach adjusts to non-\gls{IID} data distributions by adaptively
training a \gls{MoE} for clients that share similar data distributions.%
We explore a wide spectrum of data distribution
settings: ranging from the same distribution for all clients, all the way to
different distributions for each client. Our aim
is an end-to-end framework that performs comparable or better than
vanilla \gls{FL} and is \emph{robust} in all of these 
 settings.

In order to achieve personalization, the authors 
of~\cite{DBLP:conf/nips/GhoshCYR20} introduce a method for training cluster models using \gls{FL}.
We show that their solution does not perform well in our settings, where only one or a few of
the cluster models converge. To solve this, inspired by the \gls{MAB} field, 
we employ an efficient and effective way of
balancing exploration and exploitation of these cluster models.
As proposed by the authors 
of~\cite{peterson2019private,DBLP:journals/corr/abs-2010-02056}, we add a local model
and use a \gls{MoE} that learns to weigh, and make use of, all of the available models to 
produce a better personalized inference, see~\cref{fig:overview}.

In summary, our main contributions are:%
\begin{enumerate}[itemsep=1pt, topsep=0pt]
    \item We devise an \gls{FL} algorithm which improve 
    upon~\cite{DBLP:conf/nips/GhoshCYR20} by balancing exploration and
    exploitation to produce better adapted cluster models,
    see~\cref{sec:cl-fl-moe};
    \item We use said cluster models as expert models in an \gls{MoE}
    to improve performance, described in~\cref{sec:cl-fl-moe};
    \item An extensive 
    analysis\footnote{The source code for the experiments can be found 
    at \url{https://github.com/EricssonResearch/fl-moe}.} of our approach with respect to different non-IID aspects
    that also considers the distribution of client performance, see~\cref{sec:results}.
\end{enumerate}%
\section{Background}
\subsection{Problem formulation}
Consider a distributed and decentralized \gls{ML} setting with~\gls{K}
clients. Each client \({\gls{k} \in \left\{1, 2, \ldots, \gls{K}\right\}}\) has
access to a local data partition \gls{Pk} that never leaves the client where
\({\gls{nk} =\vert\gls{Pk}\vert}\) is the number of local data samples.

In this paper we are considering a multi-class classification problem where we
have \({\gls{n} = \sum_{\gls{k}=1}^{\gls{K}} \gls{nk}}\)
data samples~\gls{xi},
indexed \({\gls{i} \in \left\{1, 2, \ldots, \gls{nk}\right\}}\), and output class label~\gls{yi}
is in a finite set. We further divide each client partition \gls{Pk} into
local training and test sets. We are interested in performance on
the local test set in a non-\gls{IID} setting, see~\cref{sec:regimes_of_noniid}.

\subsection{Regimes of non-IID data}\label{sec:regimes_of_noniid} In any
decentralized setting it is common to have non-\gls{IID} data that can
be of non-identical client
distributions~\cite{DBLP:conf/icml/HsiehPMG20,DBLP:journals/corr/abs-1912-04977},
and which can be characterized as:

\begin{itemize}
\item \emph{Feature distribution skew} (covariate-shift). The feature
distributions are different between clients.
Marginal distributions \(\gls{P}\left(\gls{x}\right)\) varies,
but \(\gls{P}\left(\gls{y} \middle\vert \gls{x}\right)\) is shared;
\item \emph{Label distribution skew} (prior probability shift, or class imbalance).
    The distribution of class labels vary between clients, so
    that \(\gls{P}\left(\gls{y}\right)\) varies but \(\gls{P}\left(\gls{x} \middle\vert \gls{y} \right)\)
    is shared;
\item \emph{Same label, different features} (concept shift). The conditional distributions
\(\gls{P}\left(\gls{x} \middle\vert \gls{y}\right)\) varies between clients but
\(\gls{P}\left(\gls{y}\right)\) is shared;
\item \emph{Same features, different label} (concept shift). The conditional distribution \(\gls{P}\left(\gls{y} \middle\vert \gls{x}\right)\)
varies between clients, but \(\gls{P}\left(\gls{x}\right)\) is shared;
\item \emph{Quantity skew} (unbalancedness). Clients have different amounts of data.
\end{itemize}

Furthermore, the data independence between clients and between data samples within
a client can also be violated.

\subsection{Federated Learning}\label{sec:fl}
In a centralized \gls{ML} solution data that may be potentially privacy-sensitive is
collected to a central location. One way of improving privacy is to use a 
collaborative \gls{ML} algorithm such as 
\gls{FedAvg}~\cite{DBLP:conf/aistats/McMahanMRHA17}. In \gls{FedAvg} training 
of a global model \(\gls{fg}(\gls{x}, \gls{wg})\)
is distributed, decentralized and synchronous. A parameter server
coordinates training on many clients over several communication rounds 
until convergence.

In communication round \gls{t}, the parameter server selects a
fraction \gls{C} out of \gls{K} clients as the set~\gls{St}.
Each selected client~\({\gls{k}\in\gls{St}}\) will train locally 
on~\gls{nk} data samples~\({(\gls{xi}, \gls{yi}), \gls{i} \in \gls{Pk}}\),
for \gls{E} epochs before an update is sent to the parameter server.
The parameter server performs aggregation of all received updates and updates the
global model parameters~\gls{wg}. Finally, the new global model parameters
are distributed to all clients.

We can now define our objective as
\begin{equation}
\min_{\gls{wg} \in \R^d} \gls{flf}(\bm{w}) \overset{\Delta}{=} \min_{\gls{wg} \in \R^d} \underbracket[.12pt][5pt]{\sum_{k=1}^{\gls{K}} \frac{\gls{nk}}{n} \overbracket[.12pt][6pt]{\frac{1}{\gls{nk}} \sum_{i \in \gls{Pk}} \underbracket[.12pt][6pt]{\gls{l}\left(\gls{xi}, \gls{yi}, \gls{wg}\right)}_{\text{sample $i$ loss}}}^{\text{client $k$ average loss}}}_{\text{population average loss}},
\label{eq:loss}
\end{equation}
where \({\gls{l}\left(\gls{xi}, \gls{yi}, \gls{wg}\right)}\) is the
loss for \({\gls{yi}, \gls{yhatg} = \gls{fg}\left(\gls{xi}, \gls{wg}\right)}\).
In other words, we aim to minimize the average loss of the global model 
over all clients in the population.

\subsection{Iterative Federated Clustering}
In many real distributed use-cases, data is naturally non-\gls{IID} and clients
form clusters of \emph{similar} clients. A possible improvement
over \gls{FedAvg} is to introduce cluster models that map to these clusters,
but the problem of identifying clients that belong to these clusters remains.
We aim to find clusters, subsets of the population of clients,
that benefit more from training together within the subset, as opposed to
training with the entire population.

Using \gls{IFCA}~\cite{DBLP:conf/nips/GhoshCYR20} we set the expected 
largest number of clusters to be \gls{J} and initialize one cluster model
with weights \gls{wgj} per cluster \({j \in \left\{1,2,\ldots,\gls{J}\right\}}\).
At communication round \gls{t} each selected client~\gls{k}
performs a cluster identity estimation, where it selects the cluster
model \(\gls{jhat}^k\) that has the lowest estimated loss on the local training set.
The cluster model parameters \gls{wgj} at time \(t+1\) are then updated by using only updates
from clients the \gls{j}th selected cluster model, so that (using model 
averaging~\cite{DBLP:conf/aistats/McMahanMRHA17,DBLP:conf/nips/GhoshCYR20})
\begin{eqnarray}
\gls{nj} \leftarrow & \sum_{k \in \left\{\gls{St} \,\middle\vert\, \gls{jhat}^k = j\right\}} \gls{nk},\\
\gls{wgjt1} \leftarrow & \sum_{k \in \left\{\gls{St} \,\middle\vert\, \gls{jhat}^k = j\right\}} \frac{\gls{nk}}{\gls{nj}}\gls{wkt1}.
\end{eqnarray}

\subsection{Federated Learning using a Mixture of Experts}\label{sec:fl-moe}
In order to construct a personalized model for each 
client,~\cite{DBLP:journals/corr/abs-2010-02056} first add a local
expert model \({\gls{flk}(\gls{x}, \gls{wlk})}\) that is trained only on local
data. Recall the global model \({\gls{fg}(\gls{x}, \gls{wg})}\) from~\cref{sec:fl}.
The authors of~\cite{DBLP:journals/corr/abs-2010-02056} then 
\emph{learn to weigh} the local expert model and the global model 
using a gating function from
\gls{MoE}~\cite{DBLP:journals/neco/JacobsJNH91,peterson2019private,DBLP:journals/corr/abs-2002-05516}.
The gating function takes the same input~\gls{x} and outputs a weight for each 
of the expert models. It uses a Softmax in the output layer 
so that these weights sum to~\num{1}. 
We define \({\gls{hk}\left(\gls{x}, \gls{whk}\right)}\) as the gating
function for client~\gls{k}. 
The same model architectures are used for all local models, so
\({\gls{hk}(\gls{x}, \gls{w}) = \gls{hkp}(\gls{x}, \gls{w})}\) and
\({\gls{flk}(\gls{x}, \gls{w}) = \gls{flkp}(\gls{x}, \gls{w})}\) for all pairs of clients~\({\gls{k}, \gls{kp}}\).
For simplicity, we write
 \({\gls{fl}\left(\gls{x}\right) = \gls{flk}\left(\gls{x}, \gls{wlk}\right)}\) and
 \({\gls{h}\left(\gls{x}\right) = \gls{hk}\left(\gls{x}, \gls{whk}\right)}\) for each client~\gls{k}.
Parameters \gls{wlk} and \gls{whk} are local to client~\gls{k} and not
shared. Finally, the personalized inference is
\begin{equation}
\gls{yhath} = \gls{h}\left(\gls{x}\right) \gls{fl}\left(\gls{x}\right) + \left[1 - \gls{h}\left(\gls{x}\right)\right] \gls{fg}\left(\gls{x}\right).
\label{eq:fl-moe}
\end{equation}

\section{Adaptive Expert Models for Personalization}
\subsection{Framework overview and motivation}%
\label{sec:cl-fl-moe}

In \gls{IFCA}, after the training phase, the cluster model with the lowest loss
on the validation set is used for all future inferences. All other cluster
models are discarded in the clients. A drawback of \gls{IFCA} is therefore that
it does not use all the information available in the clients in form of unused
cluster models. Each client has access to the full set of
cluster models, and our hypothesis is that if a client can make use of
\emph{all} of these models we can increase performance.

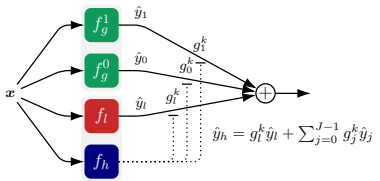
\begin{figure}[!tbh]
\centering
\begin{adjustbox}{width=.6\linewidth}%
    \def\clusters{2}
\tikzset{%
  >={Latex[width=2mm,length=3mm]},
  model/.style = {rectangle, rounded corners,
                           minimum width=.75cm, minimum height=.75cm,
                           text width=.75cm,
                           text centered, font=\large\sffamily, fill=EricssonGray4,
                           inner sep=0pt, outer sep=0pt},
  local/.style = {model, fill=EricssonRed4, text=EricssonWhite},
  federated/.style = {model, fill=EricssonGreen1, text=EricssonWhite},
  gate/.style = {model, fill=EricssonBlue1, text=EricssonWhite},
  line/.style = {thick, rounded corners, font=\large},
  connector/.style={
                font=\large
            },
            rectangle connector/.style={
                rounded corners=.25cm,
                connector,
                to path={(\tikztostart) -- ++(#1,0pt) \tikztonodes |- (\tikztotarget) },
                pos=0.5
            },
            rectangle connector/.default=-2cm,
            straight connector/.style={
                connector,
                to path=--(\tikztotarget) \tikztonodes
            },
    every node/.style={font={\small}},
    equation/.style={font={\normalsize}}
}

\begin{tikzpicture}


  \pgfmathsetmacro\midy{(\clusters-3) / 2}
  \pgfmathsetmacro\xoffset{2+\clusters*\gatesep}
  \pgfmathsetmacro\highestk{\clusters - 1}

  \coordinate (federated) at (-0.55cm,\highestk+0.45);
  \coordinate (client) at (-0.55cm,-2.45cm);

  \path [fill=EricssonGray5, rounded corners] (-1.45cm,-0.45cm) rectangle (federated);
  \path [fill=EricssonGray5, rounded corners] (-1.45cm,-0.55cm) rectangle (client);


  \node (input) at (-3cm, \midy) {$\bm{x}$};

  \node [gate] (gate) at (-1cm,-2cm) {\gls{h}};

  \node (local) [local] at (-1cm, -1cm) {\gls{fl}};
  \draw [->, line] (input) -- ($(local.west) + (-.5cm,0cm)$) -- (local.west);
  \draw [->, line] (input) -- ($(gate.west) + (-.5cm,0cm)$) -- (gate.west);

  \node (sum) [circle, inner sep=0pt, outer sep=0pt, draw=black, thick] at (\xoffset, \midy) {\suma};
  \node [right of=sum, right] (output) {};
  \draw [->, line] (sum) -- (output) node [equation, midway, below=.5cm] {$\gls{yhath} = \gls{glk}\gls{yhatl} + \sum_{\gls{j}=0}^{\gls{J}-1} \gls{gjk}\gls{yhatj} $};

  \foreach \y in {0,...,\highestk} {
        \node (cluster_\y) [federated] at (-1cm, \y) {$f^{\y}_g$};
        \draw [->, line] (input) -- ($(cluster_\y.west) + (-.5cm,0cm)$) -- (cluster_\y.west);
        \draw [->, line] (cluster_\y.east) -- ($(cluster_\y.east) + (0.5cm, 0cm)$) -- (sum) node[pos=0,above] {$\hat{y}_\y$};

        \coordinate (a) at ($(gate.east) + (1.5+ \gatesep*\y, 0)$);
        \coordinate (b) at ($(a) + (0,5)$);
        \coordinate (d) at ($(cluster_\y.east) + (0.5cm, 0cm)$);
        \coordinate (c) at (intersection of a--b and d--sum);

        \draw [dotted, -|, shorten >=0.1cm, line ] (gate.east) -- (a) -- (c) node[above] {$g^k_\y$};
  }

  \draw [->, line] (local.east) -- ($(local.east) + (0.5cm, 0cm)$) --  (sum) node[pos=0,above] {\gls{yhatl}};
  \coordinate (a) at ($(gate.east) + (1.5 + \gatesep*-1, 0)$);
  \coordinate (b) at ($(a) + (0,5)$);
  \coordinate (d) at ($(local.east) + (0.5cm, 0cm)$);
  \coordinate (c) at (intersection of a--b and d--sum);
  \draw [dotted, -|, shorten >=0.1cm, line] (gate.east) -- (a) -- (c) node[above] {\gls{glk}};

\end{tikzpicture}%
\end{adjustbox}%
\caption{Our solution with \num{2} global cluster models. Each client~\gls{k} has one local expert model
\(\gls{fl}(\gls{x},\gls{wlk})\) and share \({\gls{J}=2}\) expert cluster
models \(\gls{fgj}(\gls{x}, \gls{wgj})\) with all other clients.
A gating
model \(\gls{h}(\gls{x},\gls{whk})\) is used to weigh the expert cluster models
and produce a personalized inference~\gls{yhath} from the input~\gls{x}.}%
\label{fig:cl-moe}%
\end{figure}

It is sometimes advantageous to incorporate a local model,
as in \Cref{sec:fl-moe}, especially when the local data distribution is
very different from other clients. We therefore modify
the \gls{MoE}~\cite{DBLP:journals/corr/abs-2010-02056} to
incorporate \emph{all} the cluster models from~\gls{IFCA}~\cite{DBLP:conf/nips/GhoshCYR20} 
\emph{and} the local model as expert models in the mixture,
see~\cref{fig:cl-moe}. We revise~\eqref{eq:fl-moe} to%
\begin{equation}
\gls{yhath} = \gls{gl}\gls{flk}\left(\bm{x}\right) + \sum_{j=0}^{\gls{J}-1} \gls{gjk}\gls{fgj}\left(\bm{x}\right),
\label{eq:cl-fl-moe}
\end{equation} 
where \(\gls{gl}\) is the local model expert weight, and
\(\gls{gjk}\) is the cluster model expert weight for
cluster~\gls{j} from \(\gls{hk}\left(\gls{x}\right)\), see~\cref{fig:cl-moe}.

However, importantly, we note that
setting~\gls{J} in~\cite{DBLP:conf/nips/GhoshCYR20} to a large value produces
few cluster models that actually converge, which lowers
performance when used in a \gls{MoE}. We differ
from~\cite{DBLP:conf/nips/GhoshCYR20} in the cluster estimation step in
that we select the same number of clients
\({\gls{Ks} = \left\lceil \gls{C}\gls{K} \right\rceil}\) in
every communication round, regardless of \gls{J}. This spreads out more evenly
over the global cluster models. Since cluster models are
randomly initialized we can end up updating one cluster model more than the
others by chance. In following communication rounds, a client is more likely
to select this cluster model, purely because it has been updated more. This 
also has the effect that as \gls{J} increases, the quality of the updates 
are reduced as they are averaged from a smaller set of clients. In turn, this
means that we needed more iterations to converge.
Therefore, we make use of the
\gls{eps}-greedy algorithm~\cite{DBLP:conf/nips/Sutton95} in order to allow each
client to prioritize gathering information (\emph{exploration}) of the 
cluster models or use the estimated best
cluster model (\emph{exploitation}). In each iteration we select a random cluster model with
probability~\gls{eps} and the currently best otherwise, see~\cref{alg:fl.cluster}.

By using the \gls{eps}-greedy algorithm we make more expert 
models converge. We can then use the gating
function~\gls{hk} from \gls{MoE} to adapt to the underlying data
distributions and weigh the different expert models. We outline our setup
in~\cref{fig:overview} and provide details in~\cref{fig:cl-moe,alg:fl.server,alg:fl.update,alg:fl.client,alg:fl.cluster}.

When a cluster model has converged it is not cost-effective to transmit
this cluster model to every client, so by using per-model early stopping we can reduce
communication in both uplink and downlink. Specifically, before training we
initialize~\({\gls{Jremset}=\gls{Jset}}\). When early stopping is
triggered for a cluster model we remove that cluster model from the
set~\gls{Jremset}. The early-stopping algorithm is described
in~\cref{alg:fl.server}.

\begin{algorithm*}[tb]
\begin{algorithmic}[1]
\Procedure{server}{$\gls{C},\gls{K}$}
    \State$\text{initialize}\,\gls{Jremset} \gets \gls{Jset}, \left\{\gls{wgjt0} \,\middle|\, j \in \gls{Jremset}\right\}$ \Comment{Initialize \gls{J} global cluster models}
    \State$\gls{Ks} \gets \left\lceil \gls{C}\gls{K} \right\rceil$ \Comment{Number of clients to select per communication round}
    \For{$\gls{t} \in \{1,2, \ldots \}$}\Comment{Until convergence}
        \State$\gls{St} \subseteq \gls{S}, \vert\gls{St}\vert = \gls{Ks}$\hfill\Comment{Random sampling of \gls{Ks} clients}
        \ForAll{$\gls{k} \in \gls{St}$}\Comment{For all clients, in parallel}
             \State$\gls{wkt1}, \gls{nk}, \gls{jhat}^k \gets k.\flconstfnt{client}\left(\left\{\gls{wgj} \,\middle|\, j \in \gls{Jremset}\right\}\right)$\Comment{Local training (\cref{alg:fl.client})}
        \EndFor%
        \ForAll{$\gls{j} \in \gls{Jset}$} \Comment{For all cluster models}
            \State$\gls{nj} \gets \sum_{k \in \left\{\gls{St} \,\middle\vert\, \gls{jhat}^k = j\right\}} \gls{nk}$\hfill\Comment{\parbox[t]{.575\linewidth}{Total number of samples for cluster model \gls{j} from clients where $\gls{j} = \gls{jhat}$}}
            \State$\gls{wgjt1} \gets \sum_{k \in \left\{\gls{St} \,\middle\vert\, \gls{jhat}^k = j\right\}} \frac{\gls{nk}}{\gls{nj}}\gls{wkt1}$\hfill\Comment{\parbox[t]{.4\linewidth}{Update cluster model \gls{j} with clients where $\gls{j} = \gls{jhat}$}}
            \If{Early stopping triggered for model \gls{j}}
                \State$\gls{Jremset} \gets \gls{Jremset} \setminus \gls{j}$\Comment{\emph{Optional:} Remove \gls{j} from the set of selectable cluster models \gls{Jremset}}
            \EndIf%
        \EndFor%
    \EndFor%
\EndProcedure%
\algstore{bkbreak1}
\end{algorithmic}
\caption{Adaptive Expert Models for \gls{FL} --- server}\label{alg:fl.server}
\end{algorithm*}%
\begin{algorithm*}[tb]
\begin{algorithmic}[1]
\algrestore{bkbreak1}
\Procedure{client}{$\left\{\gls{wgj} \,\middle|\,j \in \gls{Jremset}\right\}$}
    \State$\gls{jhat} \gets \flconstfnt{cl.-est.}\left(\varepsilon, \left\{\gls{wgj} \,\middle|\, j \in \gls{Jremset}\right\}\right),\, \gls{nk} \gets \vert\gls{Pk}\vert$\Comment{Estimate cluster belonging (\cref{alg:fl.cluster})}
    \State$\gls{wkt1} \gets \flconstfnt{update}\left(\gls{wgjht}, \gls{nk}\right)$\Comment{Perform local training using cluster model \gls{jhat} (\cref{alg:fl.update})}
    \State\textbf{return} $(\gls{wkt1}, \gls{nk}, \gls{jhat})$
\EndProcedure%
\algstore{bkbreak2}
\end{algorithmic}
\caption{Adaptive Expert Models for \gls{FL} --- client}\label{alg:fl.client}
\end{algorithm*}%
\begin{algorithm*}[tb]
\begin{algorithmic}[1]
\algrestore{bkbreak2}
\Procedure{cl.-est.}{$\varepsilon, \left\{\gls{wgj} \,\middle|\, j \in \gls{Jremset}\right\}$}
    \State\textbf{return} $\gls{jhat} \gets
    \begin{cases}
        \argminB_{j \in \gls{Jremset}} \sum_{i \in \gls{Pk}} \gls{l}\left(\gls{xi}, \gls{yi}, \gls{wgj}\right) & \text{with prob.} \; 1 - \varepsilon \hspace{8em}\text{\Comment{Lowest loss cluster model}}\\
        \mathcal{U}\left\{1,J\right\} & \text{with prob.} \; \varepsilon \hfill\text{\Comment{Random assignment}}
    \end{cases}$%
\EndProcedure%
\algstore{bkbreak3}
\end{algorithmic}
\caption{Adaptive Expert Models for \gls{FL} --- cluster assignment}\label{alg:fl.cluster}
\end{algorithm*}%
\begin{algorithm*}[tb]
\begin{algorithmic}[1]
\algrestore{bkbreak3}
\Procedure{update}{$\gls{wkt1}, \gls{nk}$}\Comment{Mini-batch gradient descent}
    \For{$e \in \gls{Eset}$}\Comment{For a few epochs}
        \ForAll{batches of size $B$}\Comment{Batch update}
            \State$\gls{wkt1} \gets \gls{wkt1} - \frac{\gls{eta}}{\gls{B}} \nabla_{\gls{wkt1}} \sum_{i=1}^{\gls{B}} l_i\left(\gls{xi}, \gls{yi}, \gls{wkt1}\right)$\Comment{Local parameter update}
        \EndFor%
    \EndFor%
    \State\textbf{return} $\gls{wkt1}$
\EndProcedure%
\end{algorithmic}
\caption{Adaptive Expert Models for \gls{FL} --- local update}\label{alg:fl.update}
\end{algorithm*}
\section{Experiments}\label{sec:experiments}

\subsection{Datasets}
\glsunset{CIFAR}

We use three different
datasets, with different non-\gls{IID} characteristics, 
in which the task is an image multi-class classification task
with varying number of classes. 
\begin{itemize}
    \item \textbf{\cifar{}}~\cite{Krizhevsky2009}, where we use a
     technique from~\cite{DBLP:journals/corr/abs-2010-02056} to create client
     partitions with a controlled \emph{Label distribution skew},
     see~\cref{sec:noniidsampling};
    \item \textbf{Rotated \cifar{}}~\cite{DBLP:conf/nips/GhoshCYR20}, where
     the client feature distributions are
     controlled by rotating \cifar{} images --- an example
     of \emph{same label, different features};
    \item \textbf{\gls{FEMNIST}}~\cite{DBLP:journals/corr/abs-1812-01097,Cohen2017}
     with handwritten
     characters written by many writers,
     exhibiting many of the non-\gls{IID} characteristics outlined
     in~\cref{sec:regimes_of_noniid}.
\end{itemize}

\subsection{Non-IID sampling}\label{sec:noniidsampling}

In order to construct a
non-\gls{IID} dataset from the \cifar{} dataset~\cite{Krizhevsky2009} with the
properties of class imbalance that we are interested in we first look
at~\cite{DBLP:conf/aistats/McMahanMRHA17}. A \emph{pathological non-IID}
dataset is constructed by sorting the dataset by label, dividing it into shards
of \SI{300}{\samples} and giving each client \num{2}~shards.

However, as in~\cite{DBLP:journals/corr/abs-2010-02056}, we are interested in varying the
degree of \emph{non-IIDness} and therefore we assign two majority classes to
each client which make up a fraction \gls{p} of the data samples of the client.
The remainder fraction \((1-p)\) is sampled uniformly from the other \num{8}~classes.
When \(p=0.2\) each class has an equal probability of being sampled. A similar
case to the \emph{pathological non-IID} above is represented by~\(p=1\).
In reality, \gls{p} is unknown.

\subsection{Model architecture}

We start with the benchmark model defined
in~\cite{DBLP:journals/corr/abs-1812-01097} which is a \gls{CNN} model with two
convolutional layers and one fully connected layer with fixed hyperparameters.
However, in our case where \(\gls{nk}\) is small, the
local model is prone to over-fitting, so it is desirable to have a
model with lower capacity. Similarly, the gating model is also prone to
overfitting due to both a small local dataset and the fact that it aims to solve
a multi-label classification problem with fewer classes (expert models),
than in the original multi-class classification problem. The local model,
gating model and cluster models therefore share the
same underlying architecture, but have hyperparameter individually tuned,
see \cref{sec:hptuning}. The
\texttt{AdamW}~\cite{DBLP:conf/iclr/LoshchilovH19} optimizer is used to
train the local model and the gating model, while
\gls{SGD}~\cite{DBLP:journals/siamrev/BottouCN18} is used to train the cluster
models to
avoid issues related to momentum parameters when averaging. We use negative
log-likelihood loss in~\eqref{eq:loss}.

\subsection{Hyperparameter tuning}\label{sec:hptuning}
Hyperparameters are tuned using~\cite{DBLP:journals/corr/abs-1807-05118} in
four stages. For each model we tune the learning rate~\gls{eta}, the number of
filters in two convolutional layers,
the number of hidden units in the fully connected layer, dropout, and weight
decay. For the \gls{eps}-greedy exploration method we also tune~\gls{eps}.

First, we tune the hyperparameters for a local model and for a single
global model. Thereafter, we tune the hyperparameters for the gating model
using the best hyperparameters found in the
earlier steps. Lastly, we tune
\gls{eps} with two cluster models~\({\gls{J}=2}\). For the no exploration
experiments we set~\({\gls{eps}=0}\).

Hyperparameters depend on~\gls{p}
and~\gls{J} but we tune the hyperparameters for a fixed majority class
fraction \(\gls{p}=0.2\), which corresponds to the \gls{IID} case. The tuned
hyperparameters are then used for all experiments. We show
that our method is still robust in the fully non-\gls{IID} case
when~\({\gls{p}=1}\). See~\cref{tbl:paramscifar} for tuned hyperparameters
in the \cifar{} experiment.

\begin{table*}[tb]
\centering
\begin{small}
\begin{tabular}{lS[table-format = +1.2e+2,tight-spacing=true,retain-zero-exponent=true,scientific-notation=true]cccS[table-format = 2.2]S[table-format = +1.2e+2,tight-spacing=true,retain-zero-exponent=true,scientific-notation=true]cS[table-format = 2.2]}
\toprule
\textbf{Model} &   $\eta$ & \textbf{Conv1} & \textbf{Conv2} & \textbf{FC} &  \textbf{Dropout} &  \textbf{Weight Dec.} & \textbf{E} &  $\varepsilon$ \\
\midrule
        Global & 0.005857 &            128 &             32 &        1024 &          0.799598 &              0.001097 &          3 &       0.333253 \\
         Local & 0.000269 &             32 &            256 &         256 &          0.762102 &              0.009886 &            &                \\
          Gate & 0.000003 &             12 &             12 &           8 &          0.775030 &              0.000688 &            &                \\
\bottomrule
\end{tabular}

\end{small}
\caption{Tuned hyper-parameters in the \cifar{} experiment for the global
cluster models, the local models and the gating model.}\label{tbl:paramscifar}
\end{table*}

\subsection{Results}\label{sec:results}
We summarize our results for the class imbalance case exemplified with the
\cifar{} dataset in~\cref{tbl:results_J6}. In~\cref{fig:exp_strat_3}, we see
an example of how the performance varies when we increase the non-\gls{IID}-ness
factor \gls{p} for the case when \(\gls{J}=3\).
In~\cref{fig:baseline_exp_strat_3} we see the performance of
\gls{IFCA}~\cite{DBLP:conf/nips/GhoshCYR20} compared to our
solution in \cref{fig:epsgreedy_exp_strat_3}. We also compare to: a local model
fine-tuned from the best cluster model, an entirely local model, and
an ensemble model where we include \emph{all} cluster models as well as the
local model with equal weights.
In~\cref{fig:cifar10_methods_1} we vary the number of cluster models \gls{J} for
different values of the majority class fraction~\gls{p}.

\begin{table*}[ptb]
\centering
\begin{small}
\begin{tabular}{c@{\hskip 16pt}lc@{\hskip 16pt}S[table-format = 2.2]S[table-format = 2.2]@{\hskip 16pt}S[table-format = 2.2]S[table-format = 2.2]@{\hskip 16pt}S[table-format = 2.2]S[table-format = 2.2]@{\hskip 16pt}S[table-format = 2.2]S[table-format = 2.2]@{\hskip 16pt}S[table-format = 2.2]S[table-format = 2.2]}
\toprule
&    &                & \multicolumn{2}{c}{\textbf{MoE}} & \multicolumn{2}{c}{\textbf{IFCA}} & \multicolumn{2}{c}{\textbf{Ensemble}} & \multicolumn{2}{c}{\textbf{Fine-tuned}} & \multicolumn{2}{c}{\textbf{Local}} \\%
\cmidrule(lr){4-5}\cmidrule(lr){6-7}\cmidrule(lr){8-9}\cmidrule(lr){10-11}\cmidrule(lr){12-13}%
$p$&\textbf{Exp.\ strategy}& \# trials &                                $\mu$ &                                 $\sigma$ &                                     $\mu$ &                                  $\sigma$ &             $\mu$ &  $\sigma$ &               $\mu$ &  $\sigma$ &          $\mu$ &  $\sigma$ \\
\midrule
\multirow{2}{*}{0.2} & $\varepsilon{}$-greedy (ours) &         7 &  \color{EricssonBlue1}\textbf{72.39} &  \color{EricssonBlue1}1.2644215692410141 &                                     70.38 &                                   0.74153 &         70.817143 &  1.866099 &               70.16 &  0.614709 &      38.520000 &  0.752330 \\
    & No exploration &         6 &                            57.734286 &                                 1.951776 &   \color{EricssonGreen1}71.25428571428571 &  \color{EricssonGreen1}0.8595901570488325 &         58.577143 &  1.958654 &           70.125714 &  1.040430 &      38.062857 &  0.873570 \\
\hdashline%
\multirow{2}{*}{0.4} & $\varepsilon{}$-greedy (ours) &         6 &  \color{EricssonBlue1}\textbf{72.05} &  \color{EricssonBlue1}1.7896033079987363 &                                     68.59 &                                  1.003015 &         69.963333 &  2.333355 &           70.283333 &  1.054546 &      43.360000 &  0.470574 \\
    & No exploration &         9 &                              60.1175 &                                 2.369724 &              \color{EricssonGreen1}68.295 &  \color{EricssonGreen1}1.5611259672794806 &         59.535000 &  1.312609 &               69.42 &  1.543317 &      43.162500 &  0.708676 \\
\hdashline%
\multirow{2}{*}{0.6} & $\varepsilon{}$-greedy (ours) &         8 &  \color{EricssonBlue1}\textbf{75.22} &  \color{EricssonBlue1}0.7457129921901847 &                                 66.530303 &                                  0.978719 &         71.441919 &  1.171366 &           72.502525 &  0.544806 &      54.633838 &  0.329120 \\
    & No exploration &         9 &                            67.936026 &                                 0.749755 &  \color{EricssonGreen1}61.468013468013474 &  \color{EricssonGreen1}2.2700335701522274 &         65.148148 &  0.870650 &           68.272727 &  1.712328 &      55.037037 &  0.449134 \\
\hdashline%
\multirow{2}{*}{0.8} & $\varepsilon{}$-greedy (ours) &        14 &  \color{EricssonBlue1}\textbf{81.09} &  \color{EricssonBlue1}1.1766329096823729 &                                 65.234848 &                                  2.397409 &         74.439394 &  0.756950 &           80.126263 &  1.019738 &      69.489899 &  0.575644 \\
    & No exploration &        15 &                            75.041323 &                                 0.929485 &   \color{EricssonGreen1}62.58953168044077 &  \color{EricssonGreen1}1.9468044066523198 &         70.820937 &  1.560592 &           76.488522 &  1.265099 &      69.550046 &  0.702327 \\
\hdashline%
\multirow{2}{*}{1.0} & $\varepsilon{}$-greedy (ours) &        14 &          \color{EricssonBlue1}90.755 &  \color{EricssonBlue1}0.8213403679352446 &                                   48.7925 &                                  5.352339 &         71.060000 &  4.033885 &               90.26 &  1.016493 &      86.652500 &  0.389129 \\
    & No exploration &         6 &                            88.791429 &                                 0.515733 &   \color{EricssonGreen1}60.97428571428571 &  \color{EricssonGreen1}2.0659691603285304 &         71.562857 &  9.124549 &      \textbf{91.11} &  0.334037 &      86.365714 &  0.314476 \\
\bottomrule
\end{tabular}

\end{small}
\caption{Results for \cifar{}
and~\(\gls{p} \in \{0.2, 0.4, \ldots, 1\}\) when \(\gls{J}=6\).
Mean \(\mu\) and standard deviation \(\sigma\) for our exploration method \gls{eps}-greedy and
without exploration. We compare {\color{EricssonBlue1}our proposed \gls{MoE} solution} to the
 {\color{EricssonGreen1}baseline} from \gls{IFCA}~\protect\cite{DBLP:conf/nips/GhoshCYR20}.
Our proposed solution is superior in all but one case, indicated by \textbf{bold} numbers.
 }\label{tbl:results_J6}%
\squeezeup%
\end{table*}

\begin{figure*}[!ht]
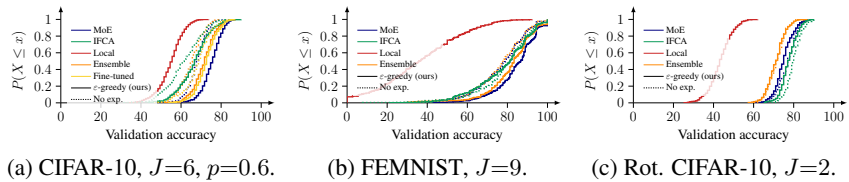

\centering
\subfloat[CIFAR-10, \({\gls{J}{=}6}\),~\({\gls{p}{=}0.6}\).\label{fig:cifar10_cdfs}]{\begin{adjustbox}{width=0.2\linewidth}\input{figures/plot_cdf_cifar10.tex}\end{adjustbox}}
\hspace{.25cm}%
\subfloat[\gls{FEMNIST}, \({\gls{J}{=}9}\).\label{fig:femnist_cdfs}]{\begin{adjustbox}{width=0.2\linewidth}\input{figures/plot_cdf_femnist.tex}\end{adjustbox}}
\hspace{.25cm}%
\subfloat[Rot. CIFAR-10, \({\gls{J}{=}2}\).\label{fig:cifar10rot_cdfs}]{\begin{adjustbox}{width=0.2\linewidth}\input{figures/plot_cdf_cifar10rot.tex}\end{adjustbox}}
\caption{\glsxtrshort{CDF} of client accuracy. Comparison between no exploration
(colored dashed lines) and
the \gls{eps}-greedy exploration method (colored solid lines).
{\color{EricssonBlue1}Our proposed \gls{MoE} solution} with \gls{eps}-greedy
exploration improves accuracy
and fairness for two of the datasets.}\label{fig:cdfs}%
\squeezeup%
\end{figure*}

An often overlooked aspect of performance in \gls{FL} is the variance between
clients. We achieve a smaller inter-client variance, shown
for \cifar{} in~\cref{fig:cifar10_cdfs,tbl:results_J6}.

We see that for \cifar{} our \gls{eps}-greedy exploration method achieves better
results for lower values of \gls{p} by allowing more of the cluster
models to converge --- thereby more cluster models are useful as
experts in the \gls{MoE}, even though the models are similar,
see~\cref{fig:cifar10_cluster_02}. For higher values of \gls{p} we see that
the cluster models are adapting to existing clusters in the data,
see~\cref{fig:cifar10_cluster_1}. The most
interesting result is seen in between these extremes,
see~\cref{fig:cifar10_cluster_06}. We note that the same
number of clients pick each cluster model as in \gls{IFCA}, but we
manage to make a better selection and achieve higher performance.
\newcommand\figpwidth{.38}
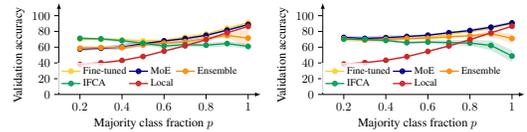
\begin{figure}[!ht]
    \centering
    \subfloat[No exploration.\label{fig:baseline_exp_strat_3}]{\begin{adjustbox}{width=\figpwidth\linewidth}
\begin{tikzpicture}

\definecolor{color0}{rgb}{0.980392156862745,0.823529411764706,0.176470588235294}
\definecolor{color1}{rgb}{1,0.549019607843137,0.0392156862745098}
\definecolor{color2}{rgb}{0.0470588235294118,0.607843137254902,0.356862745098039}
\definecolor{color3}{rgb}{0.8,0.156862745098039,0.156862745098039}

\begin{axis}[
legend cell align={left},
legend columns=2,
legend style={
  fill opacity=0.8,
  draw opacity=1,
  text opacity=1,
  at={(0.97,0.03)},
  anchor=south east,
  draw=none
},
tick align=outside,
tick pos=left,
x grid style={white!69.0196078431373!black},
xlabel={Majority class
fraction \gls{p}},
xmin=0.1, xmax=1,
xtick style={color=black},
y grid style={white!69.0196078431373!black},
ylabel={Validation accuracy},
ymin=0, ymax=100,
ytick style={color=black}
,
legend cell align={left},
legend style={nodes={scale=0.9, transform shape}},
legend image post style={mark=*},
legend columns=3
]

\path [draw=color0, fill=color0, opacity=0.2]
(axis cs:0.2,71.27)
--(axis cs:0.2,68.616)
--(axis cs:0.3,69.595)
--(axis cs:0.4,67.169)
--(axis cs:0.5,66.272)
--(axis cs:0.6,66.3989898989899)
--(axis cs:0.7,71.1424242424243)
--(axis cs:0.8,74.8787878787879)
--(axis cs:0.9,83.3080808080808)
--(axis cs:1,90.644)
--(axis cs:1,91.49)
--(axis cs:1,91.49)
--(axis cs:0.9,84.5151515151515)
--(axis cs:0.8,78.1818181818182)
--(axis cs:0.7,74.5111111111111)
--(axis cs:0.6,70.6111111111111)
--(axis cs:0.5,69.368)
--(axis cs:0.4,71.316)
--(axis cs:0.3,70.475)
--(axis cs:0.2,71.27)
--cycle;

\path [draw=blue!50.9803921568627!black, fill=blue!50.9803921568627!black, opacity=0.2]
(axis cs:0.2,60.468)
--(axis cs:0.2,55.476)
--(axis cs:0.3,57.695)
--(axis cs:0.4,57.285)
--(axis cs:0.5,62.966)
--(axis cs:0.6,67.0808065986633)
--(axis cs:0.7,68.6696959724426)
--(axis cs:0.8,73.6565656661987)
--(axis cs:0.9,81.0909097862244)
--(axis cs:1,88.072)
--(axis cs:1,89.4)
--(axis cs:1,89.4)
--(axis cs:0.9,82.4242432022095)
--(axis cs:0.8,76.2424247741699)
--(axis cs:0.7,72.1999993247986)
--(axis cs:0.6,68.8636351394653)
--(axis cs:0.5,66.48)
--(axis cs:0.4,63.581)
--(axis cs:0.3,59.595)
--(axis cs:0.2,60.468)
--cycle;

\path [draw=color1, fill=color1, opacity=0.2]
(axis cs:0.2,61.412)
--(axis cs:0.2,56.536)
--(axis cs:0.3,58.76)
--(axis cs:0.4,57.719)
--(axis cs:0.5,61.234)
--(axis cs:0.6,64.1060606060606)
--(axis cs:0.7,64.7969696969697)
--(axis cs:0.8,69.0505050505051)
--(axis cs:0.9,72.979797979798)
--(axis cs:1,59.518)
--(axis cs:1,81.736)
--(axis cs:1,81.736)
--(axis cs:0.9,76.2626262626263)
--(axis cs:0.8,73.3434343434344)
--(axis cs:0.7,68.1252525252525)
--(axis cs:0.6,66.1262626262626)
--(axis cs:0.5,64.648)
--(axis cs:0.4,61.105)
--(axis cs:0.3,60.22)
--(axis cs:0.2,61.412)
--cycle;

\path [draw=color2, fill=color2, opacity=0.2]
(axis cs:0.2,72.496)
--(axis cs:0.2,70.364)
--(axis cs:0.3,69.925)
--(axis cs:0.4,66.207)
--(axis cs:0.5,62.784)
--(axis cs:0.6,59.4292929292929)
--(axis cs:0.7,61.159595959596)
--(axis cs:0.8,60.1919191919192)
--(axis cs:0.9,61.7626262626263)
--(axis cs:1,58.734)
--(axis cs:1,63.594)
--(axis cs:1,63.594)
--(axis cs:0.9,67.1767676767677)
--(axis cs:0.8,65.0606060606061)
--(axis cs:0.7,66.6141414141414)
--(axis cs:0.6,64.6313131313131)
--(axis cs:0.5,66.156)
--(axis cs:0.4,70.126)
--(axis cs:0.3,71.28)
--(axis cs:0.2,72.496)
--cycle;

\path [draw=color3, fill=color3, opacity=0.2]
(axis cs:0.2,39.044)
--(axis cs:0.2,36.866)
--(axis cs:0.3,38.98)
--(axis cs:0.4,42.165)
--(axis cs:0.5,47.544)
--(axis cs:0.6,54.3838383838384)
--(axis cs:0.7,60.9454545454545)
--(axis cs:0.8,68.6666666666667)
--(axis cs:0.9,77.4444444444444)
--(axis cs:1,85.894)
--(axis cs:1,86.642)
--(axis cs:1,86.642)
--(axis cs:0.9,78.9949494949495)
--(axis cs:0.8,70.5454545454546)
--(axis cs:0.7,62.3505050505051)
--(axis cs:0.6,55.4343434343434)
--(axis cs:0.5,48.73)
--(axis cs:0.4,43.89)
--(axis cs:0.3,40.935)
--(axis cs:0.2,39.044)
--cycle;

\addplot [very thick, color0, mark=*, mark size=2, mark options={solid}]
table {%
0.2 70.1257142857143
0.3 69.98
0.4 69.42
0.5 68.0742857142857
0.6 68.2727272727273
0.7 72.6237373737374
0.8 76.4885215794307
0.9 83.8686868686869
1 91.1114285714286
};
\addlegendentry{Fine-tuned}
\addplot [very thick, blue!50.9803921568627!black, mark=*, mark size=2, mark options={solid}]
table {%
0.2 57.7342857142857
0.3 58.6666666666667
0.4 60.1175
0.5 64.3514285714286
0.6 67.9360255940755
0.7 70.914140625
0.8 75.0413225416704
0.9 81.5925931294759
1 88.7914285714286
};
\addlegendentry{MoE}
\addplot [very thick, color1, mark=*, mark size=2, mark options={solid}]
table {%
0.2 58.5771428571429
0.3 59.5433333333333
0.4 59.535
0.5 62.72
0.6 65.1481481481482
0.7 67.0656565656566
0.8 70.8209366391184
0.9 74.8518518518518
1 71.5628571428572
};
\addlegendentry{Ensemble}
\addplot [very thick, color2, mark=*, mark size=2, mark options={solid}]
table {%
0.2 71.2542857142857
0.3 70.5633333333333
0.4 68.295
0.5 65.0742857142857
0.6 61.4680134680135
0.7 63.2121212121212
0.8 62.5895316804408
0.9 64.6296296296296
1 60.9742857142857
};
\addlegendentry{\gls{IFCA}}
\addplot [very thick, color3, mark=*, mark size=2, mark options={solid}]
table {%
0.2 38.0628571428571
0.3 40.1066666666667
0.4 43.1625
0.5 48.1942857142857
0.6 55.037037037037
0.7 61.7449494949495
0.8 69.5500459136823
0.9 78.3434343434344
1 86.3657142857143
};
\addlegendentry{Local}
\end{axis}

\end{tikzpicture}\end{adjustbox}}
    \hspace{.25cm}%
    \subfloat[\gls{eps}-greedy exploration.\label{fig:epsgreedy_exp_strat_3}]{\begin{adjustbox}{width=\figpwidth\linewidth}
\begin{tikzpicture}

\definecolor{color0}{rgb}{0.980392156862745,0.823529411764706,0.176470588235294}
\definecolor{color1}{rgb}{1,0.549019607843137,0.0392156862745098}
\definecolor{color2}{rgb}{0.0470588235294118,0.607843137254902,0.356862745098039}
\definecolor{color3}{rgb}{0.8,0.156862745098039,0.156862745098039}

\begin{axis}[
legend cell align={left},
legend columns=2,
legend style={
  fill opacity=0.8,
  draw opacity=1,
  text opacity=1,
  at={(0.97,0.03)},
  anchor=south east,
  draw=none
},
tick align=outside,
tick pos=left,
x grid style={white!69.0196078431373!black},
xlabel={Majority class
fraction \gls{p}},
xmin=0.1, xmax=1,
xtick style={color=black},
y grid style={white!69.0196078431373!black},
ylabel={Validation accuracy},
ymin=0, ymax=100,
ytick style={color=black}
,
legend cell align={left},
legend style={nodes={scale=0.9, transform shape}},
legend image post style={mark=*},
legend columns=3
]

\path [draw=color0, fill=color0, opacity=0.2]
(axis cs:0.2,71.122)
--(axis cs:0.2,69.65)
--(axis cs:0.3,68.365)
--(axis cs:0.4,69.285)
--(axis cs:0.5,69.428)
--(axis cs:0.6,71.8949494949495)
--(axis cs:0.7,75.8181818181818)
--(axis cs:0.8,79.2060606060606)
--(axis cs:0.9,82.719191919192)
--(axis cs:1,89.008)
--(axis cs:1,91.604)
--(axis cs:1,91.604)
--(axis cs:0.9,85.4555555555556)
--(axis cs:0.8,81.6606060606061)
--(axis cs:0.7,77.020202020202)
--(axis cs:0.6,73.2949494949495)
--(axis cs:0.5,70.98)
--(axis cs:0.4,71.52)
--(axis cs:0.3,71.06)
--(axis cs:0.2,71.122)
--cycle;

\path [draw=blue!50.9803921568627!black, fill=blue!50.9803921568627!black, opacity=0.2]
(axis cs:0.2,73.764)
--(axis cs:0.2,70.852)
--(axis cs:0.3,69)
--(axis cs:0.4,70.29)
--(axis cs:0.5,72.796)
--(axis cs:0.6,74.0797978477478)
--(axis cs:0.7,77.6363645744324)
--(axis cs:0.8,79.3959604759216)
--(axis cs:0.9,83.3363634948731)
--(axis cs:1,90.024)
--(axis cs:1,92.038)
--(axis cs:1,92.038)
--(axis cs:0.9,86.6141410636902)
--(axis cs:0.8,82.4424248695373)
--(axis cs:0.7,78.8989907836914)
--(axis cs:0.6,75.9070710830688)
--(axis cs:0.5,74.392)
--(axis cs:0.4,74.465)
--(axis cs:0.3,73.06)
--(axis cs:0.2,73.764)
--cycle;

\path [draw=color1, fill=color1, opacity=0.2]
(axis cs:0.2,72.654)
--(axis cs:0.2,68.214)
--(axis cs:0.3,66.295)
--(axis cs:0.4,67.425)
--(axis cs:0.5,70.284)
--(axis cs:0.6,69.9737373737374)
--(axis cs:0.7,71.8989898989899)
--(axis cs:0.8,73.4939393939394)
--(axis cs:0.9,73.7434343434344)
--(axis cs:1,65.668)
--(axis cs:1,76.17)
--(axis cs:1,76.17)
--(axis cs:0.9,80.149494949495)
--(axis cs:0.8,75.3181818181818)
--(axis cs:0.7,74.5353535353536)
--(axis cs:0.6,72.9131313131313)
--(axis cs:0.5,71.744)
--(axis cs:0.4,72.86)
--(axis cs:0.3,71.285)
--(axis cs:0.2,72.654)
--cycle;

\path [draw=color2, fill=color2, opacity=0.2]
(axis cs:0.2,71.004)
--(axis cs:0.2,69.274)
--(axis cs:0.3,68.925)
--(axis cs:0.4,67.4)
--(axis cs:0.5,64.644)
--(axis cs:0.6,65.2757575757576)
--(axis cs:0.7,63.3030303030303)
--(axis cs:0.8,62.4808080808081)
--(axis cs:0.9,56.9656565656566)
--(axis cs:1,42.073)
--(axis cs:1,55.534)
--(axis cs:1,55.534)
--(axis cs:0.9,67.8343434343435)
--(axis cs:0.8,68.5232323232323)
--(axis cs:0.7,68.2121212121212)
--(axis cs:0.6,67.8161616161616)
--(axis cs:0.5,66.712)
--(axis cs:0.4,69.875)
--(axis cs:0.3,71.215)
--(axis cs:0.2,71.004)
--cycle;

\path [draw=color3, fill=color3, opacity=0.2]
(axis cs:0.2,39.454)
--(axis cs:0.2,37.488)
--(axis cs:0.3,39.935)
--(axis cs:0.4,42.87)
--(axis cs:0.5,47.212)
--(axis cs:0.6,54.1575757575758)
--(axis cs:0.7,60.979797979798)
--(axis cs:0.8,68.8313131313131)
--(axis cs:0.9,77.6222222222222)
--(axis cs:1,86.179)
--(axis cs:1,87.205)
--(axis cs:1,87.205)
--(axis cs:0.9,78.7353535353535)
--(axis cs:0.8,70.3020202020202)
--(axis cs:0.7,62.3181818181818)
--(axis cs:0.6,54.9676767676768)
--(axis cs:0.5,48.728)
--(axis cs:0.4,44.005)
--(axis cs:0.3,40.705)
--(axis cs:0.2,39.454)
--cycle;

\addplot [very thick, color0, mark=*, mark size=2, mark options={solid}]
table {%
0.2 70.16
0.3 69.7
0.4 70.2833333333333
0.5 70.136
0.6 72.5025252525253
0.7 76.2626262626263
0.8 80.1262626262626
0.9 84.5030303030303
1 90.26
};
\addlegendentry{Fine-tuned}
\addplot [very thick, blue!50.9803921568627!black, mark=*, mark size=2, mark options={solid}]
table {%
0.2 72.3942857142857
0.3 71.1266666666667
0.4 72.05
0.5 73.552
0.6 75.2196971321106
0.7 78.3299672953288
0.8 81.0909099674225
0.9 85.3414140472412
1 90.755
};
\addlegendentry{MoE}
\addplot [very thick, color1, mark=*, mark size=2, mark options={solid}]
table {%
0.2 70.8171428571429
0.3 69.06
0.4 69.9633333333333
0.5 71.092
0.6 71.4419191919192
0.7 73.2491582491583
0.8 74.439393939394
0.9 77.1757575757576
1 71.06
};
\addlegendentry{Ensemble}
\addplot [very thick, color2, mark=*, mark size=2, mark options={solid}]
table {%
0.2 70.38
0.3 69.8566666666667
0.4 68.59
0.5 65.76
0.6 66.530303030303
0.7 65.7845117845118
0.8 65.2348484848485
0.9 62.2383838383838
1 48.7925
};
\addlegendentry{\gls{IFCA}}
\addplot [very thick, color3, mark=*, mark size=2, mark options={solid}]
table {%
0.2 38.52
0.3 40.3466666666667
0.4 43.36
0.5 48.008
0.6 54.6338383838384
0.7 61.6094276094276
0.8 69.489898989899
0.9 78.2343434343435
1 86.6525
};
\addlegendentry{Local}
\end{axis}

\end{tikzpicture}\end{adjustbox}}
\caption{Results for \cifar{}. Comparison between
no exploration
and our \gls{eps}-greedy exploration method
for \(\gls{J}=6\). {\color{EricssonBlue1}Our proposed \gls{MoE} solution} with
\gls{eps}-greedy exploration is superior in all cases from IID to pathological
non-IID class distributions, here shown by varying the
majority class fraction~\gls{p}.}%
\label{fig:exp_strat_3}%
\squeezeup%
\end{figure}

\begin{figure*}[!tbh]
    \centering
    \subfloat[\(\gls{p}=0.2\)\label{fig:k_02}]{\begin{adjustbox}{width=.18\linewidth}\input{figures/plot_k_02.tex}\end{adjustbox}}
    \hfill%
    \subfloat[\(\gls{p}=0.4\)\label{fig:k_04}]{\begin{adjustbox}{width=.18\linewidth}\input{figures/plot_k_04.tex}\end{adjustbox}}
    \hfill%
    \subfloat[\(\gls{p}=0.6\)\label{fig:k_06}]{\begin{adjustbox}{width=.18\linewidth}\input{figures/plot_k_06.tex}\end{adjustbox}}
    \hfill%
    \subfloat[\(\gls{p}=0.8\)\label{fig:k_08}]{\begin{adjustbox}{width=.18\linewidth}\input{figures/plot_k_08.tex}\end{adjustbox}}
    \hfill%
    \subfloat[\(\gls{p}=1\)\label{fig:k_1}]{\begin{adjustbox}{width=.18\linewidth}\input{figures/plot_k_1.tex}\end{adjustbox}}
\caption{Results for \cifar{}. Comparison between no exploration (colored dashed lines) and
the \gls{eps}-greedy exploration method (colored solid lines).
{\color{EricssonBlue1}Our proposed \gls{MoE} solution} with the \gls{eps}-greedy exploration outperforms all other solutions,
including the {\color{EricssonGreen1}baseline} from \gls{IFCA}~\protect\cite{DBLP:conf/nips/GhoshCYR20}.
It performs better the greater the non-IIDness, here seen by varying the
majority class fraction~\gls{p}. Furthermore, our solution is robust to changes in the number of cluster models \gls{J}.}%
\label{fig:cifar10_methods_1}%
\end{figure*}

\pgfplotsset{compat=newest,
    width=6cm,
    height=2.5cm,
    scale only axis=true,
    max space between ticks=25pt,
    try min ticks=5,
    every axis/.style={
        axis y line=left,
        axis x line=bottom,
        axis line style={thick,->,>=latex, shorten >=-.4cm},
        mark size=1,
        label style={font=\Large},
        tick label style={font=\Large}
    },
    every axis plot/.append style={thick},
    tick style={black, thick},
    colorbar style={at={(1.25, 1)}}
}
\begin{figure}[!bth]
    \centering
    \subfloat[\(\gls{p}=0.2\)\label{fig:cifar10_cluster_02}]{\begin{adjustbox}{width=0.32\linewidth}
\begin{tikzpicture}

\definecolor{color0}{rgb}{1,0.549019607843137,0.0392156862745098}

\begin{axis}[
legend cell align={left},
legend style={fill opacity=0.8, draw opacity=1, text opacity=1, draw=none},
tick align=outside,
tick pos=left,
x grid style={white!69.0196078431373!black},
xlabel={Cluster model
(sorted by frequency)},
xmin=0.375, xmax=8.625,
xtick style={color=black},
y grid style={white!69.0196078431373!black},
ylabel={Number of clients},
ymin=0, ymax=50,
ytick style={color=black}
]
\draw[draw=none,fill=blue!50.9803921568627!black] (axis cs:0.75,0) rectangle (axis cs:1,50);
\addlegendimage{ybar,ybar legend,draw=none,fill=blue!50.9803921568627!black};
\addlegendentry{No exploration}

\draw[draw=none,fill=blue!50.9803921568627!black] (axis cs:1.75,0) rectangle (axis cs:2,0);
\draw[draw=none,fill=blue!50.9803921568627!black] (axis cs:2.75,0) rectangle (axis cs:3,0);
\draw[draw=none,fill=blue!50.9803921568627!black] (axis cs:3.75,0) rectangle (axis cs:4,0);
\draw[draw=none,fill=blue!50.9803921568627!black] (axis cs:4.75,0) rectangle (axis cs:5,0);
\draw[draw=none,fill=blue!50.9803921568627!black] (axis cs:5.75,0) rectangle (axis cs:6,0);
\draw[draw=none,fill=blue!50.9803921568627!black] (axis cs:6.75,0) rectangle (axis cs:7,0);
\draw[draw=none,fill=blue!50.9803921568627!black] (axis cs:7.75,0) rectangle (axis cs:8,0);
\draw[draw=none,fill=color0] (axis cs:1,0) rectangle (axis cs:1.25,33);
\addlegendimage{ybar,ybar legend,draw=none,fill=color0};
\addlegendentry{$\varepsilon{}$-greedy (ours)}

\draw[draw=none,fill=color0] (axis cs:2,0) rectangle (axis cs:2.25,7);
\draw[draw=none,fill=color0] (axis cs:3,0) rectangle (axis cs:3.25,5);
\draw[draw=none,fill=color0] (axis cs:4,0) rectangle (axis cs:4.25,2);
\draw[draw=none,fill=color0] (axis cs:5,0) rectangle (axis cs:5.25,2);
\draw[draw=none,fill=color0] (axis cs:6,0) rectangle (axis cs:6.25,1);
\draw[draw=none,fill=color0] (axis cs:7,0) rectangle (axis cs:7.25,0);
\draw[draw=none,fill=color0] (axis cs:8,0) rectangle (axis cs:8.25,0);
\end{axis}

\end{tikzpicture}\end{adjustbox}}
    \hfill%
    \subfloat[\(\gls{p}=0.6\)\label{fig:cifar10_cluster_06}]{\begin{adjustbox}{width=0.32\linewidth}
\begin{tikzpicture}

\definecolor{color0}{rgb}{1,0.549019607843137,0.0392156862745098}

\begin{axis}[
legend cell align={left},
legend style={fill opacity=0.8, draw opacity=1, text opacity=1, draw=none},
tick align=outside,
tick pos=left,
x grid style={white!69.0196078431373!black},
xlabel={Cluster model
(sorted by frequency)},
xmin=0.375, xmax=8.625,
xtick style={color=black},
y grid style={white!69.0196078431373!black},
ylabel={Number of clients},
ymin=0, ymax=50,
ytick style={color=black}
]
\draw[draw=none,fill=blue!50.9803921568627!black] (axis cs:0.75,0) rectangle (axis cs:1,31);
\addlegendimage{ybar,ybar legend,draw=none,fill=blue!50.9803921568627!black};
\addlegendentry{No exploration}

\draw[draw=none,fill=blue!50.9803921568627!black] (axis cs:1.75,0) rectangle (axis cs:2,13);
\draw[draw=none,fill=blue!50.9803921568627!black] (axis cs:2.75,0) rectangle (axis cs:3,4);
\draw[draw=none,fill=blue!50.9803921568627!black] (axis cs:3.75,0) rectangle (axis cs:4,1);
\draw[draw=none,fill=blue!50.9803921568627!black] (axis cs:4.75,0) rectangle (axis cs:5,1);
\draw[draw=none,fill=blue!50.9803921568627!black] (axis cs:5.75,0) rectangle (axis cs:6,0);
\draw[draw=none,fill=blue!50.9803921568627!black] (axis cs:6.75,0) rectangle (axis cs:7,0);
\draw[draw=none,fill=blue!50.9803921568627!black] (axis cs:7.75,0) rectangle (axis cs:8,0);
\draw[draw=none,fill=color0] (axis cs:1,0) rectangle (axis cs:1.25,23);
\addlegendimage{ybar,ybar legend,draw=none,fill=color0};
\addlegendentry{$\varepsilon{}$-greedy (ours)}

\draw[draw=none,fill=color0] (axis cs:2,0) rectangle (axis cs:2.25,11);
\draw[draw=none,fill=color0] (axis cs:3,0) rectangle (axis cs:3.25,10);
\draw[draw=none,fill=color0] (axis cs:4,0) rectangle (axis cs:4.25,3);
\draw[draw=none,fill=color0] (axis cs:5,0) rectangle (axis cs:5.25,1);
\draw[draw=none,fill=color0] (axis cs:6,0) rectangle (axis cs:6.25,1);
\draw[draw=none,fill=color0] (axis cs:7,0) rectangle (axis cs:7.25,1);
\draw[draw=none,fill=color0] (axis cs:8,0) rectangle (axis cs:8.25,0);
\end{axis}

\end{tikzpicture}\end{adjustbox}}
    \hfill
    \subfloat[\(\gls{p}=1\)\label{fig:cifar10_cluster_1}]{\begin{adjustbox}{width=0.32\linewidth}
\begin{tikzpicture}

\definecolor{color0}{rgb}{1,0.549019607843137,0.0392156862745098}

\begin{axis}[
legend cell align={left},
legend style={fill opacity=0.8, draw opacity=1, text opacity=1, draw=none},
tick align=outside,
tick pos=left,
x grid style={white!69.0196078431373!black},
xlabel={Cluster model
(sorted by frequency)},
xmin=0.375, xmax=8.625,
xtick style={color=black},
y grid style={white!69.0196078431373!black},
ylabel={Number of clients},
ymin=0, ymax=50,
ytick style={color=black}
]
\draw[draw=none,fill=blue!50.9803921568627!black] (axis cs:0.75,0) rectangle (axis cs:1,43);
\addlegendimage{ybar,ybar legend,draw=none,fill=blue!50.9803921568627!black};
\addlegendentry{No exploration}

\draw[draw=none,fill=blue!50.9803921568627!black] (axis cs:1.75,0) rectangle (axis cs:2,1);
\draw[draw=none,fill=blue!50.9803921568627!black] (axis cs:2.75,0) rectangle (axis cs:3,1);
\draw[draw=none,fill=blue!50.9803921568627!black] (axis cs:3.75,0) rectangle (axis cs:4,1);
\draw[draw=none,fill=blue!50.9803921568627!black] (axis cs:4.75,0) rectangle (axis cs:5,1);
\draw[draw=none,fill=blue!50.9803921568627!black] (axis cs:5.75,0) rectangle (axis cs:6,1);
\draw[draw=none,fill=blue!50.9803921568627!black] (axis cs:6.75,0) rectangle (axis cs:7,1);
\draw[draw=none,fill=blue!50.9803921568627!black] (axis cs:7.75,0) rectangle (axis cs:8,1);
\draw[draw=none,fill=color0] (axis cs:1,0) rectangle (axis cs:1.25,11);
\addlegendimage{ybar,ybar legend,draw=none,fill=color0};
\addlegendentry{$\varepsilon{}$-greedy (ours)}

\draw[draw=none,fill=color0] (axis cs:2,0) rectangle (axis cs:2.25,10);
\draw[draw=none,fill=color0] (axis cs:3,0) rectangle (axis cs:3.25,10);
\draw[draw=none,fill=color0] (axis cs:4,0) rectangle (axis cs:4.25,9);
\draw[draw=none,fill=color0] (axis cs:5,0) rectangle (axis cs:5.25,4);
\draw[draw=none,fill=color0] (axis cs:6,0) rectangle (axis cs:6.25,3);
\draw[draw=none,fill=color0] (axis cs:7,0) rectangle (axis cs:7.25,2);
\draw[draw=none,fill=color0] (axis cs:8,0) rectangle (axis cs:8.25,1);
\end{axis}

\end{tikzpicture}\end{adjustbox}}
\caption{Results for \cifar{}. The number of clients
 in each cluster for the different exploration methods. Clusters are sorted,
 so that the lowest index corresponds to the most picked cluster. Our
 \gls{eps}-greedy exploration method picks the cluster models more evenly.}
\label{fig:cifar10_clusters}%
\end{figure}

For the rotated \cifar{} case we see that \gls{IFCA} manages
to assign each client to the correct clusters at \({\gls{J}=2}\), and in this
\emph{Same label, different features} case our exploration
method requires a larger~\gls{J} to achieve the same performance.
We also note the very high \({\gls{eps}=0.82}\).
More work is needed on better exploration methods for this case.

The \gls{FEMNIST} dataset represents a more difficult scenario since there are
many non-\gls{IID} aspects in this dataset. We find that for \gls{FEMNIST}
the best performance is achieved when \({\gls{J}=9}\) and
in~\cref{fig:femnist_cdfs} we show the
distribution of accuracy for the clients for the different models.
\section{Related Work}
The \gls{FedAvg} algorithm~\cite{DBLP:conf/aistats/McMahanMRHA17} is the
most prevalent algorithm for learning a global model in \gls{FL}. This
algorithm has demonstrated that an average over model
parameters is an efficient way to aggregate local models into a global model.
However, when data is non-\gls{IID}, \gls{FedAvg} converges slowly or not at 
all. This has
given rise to personalization methods for
\gls{FL}~\cite{DBLP:journals/corr/abs-1912-04977,DBLP:conf/icml/HsiehPMG20}.
Research on how to handle non-\gls{IID} data among
clients is ample and expanding. Solutions include fine-tuning
locally~\cite{DBLP:journals/corr/abs-1910-10252}, 
meta-learning~\cite{jiang2019improving,DBLP:conf/icml/FinnAL17},
\gls{MAB}~\cite{DBLP:conf/aistats/ShiSY21}, multi-task
learning~\cite{DBLP:conf/icml/00050BS21}, model heterogeneous
methods~\cite{DBLP:journals/corr/abs-2004-08546,DBLP:conf/iclr/Diao0T21},
data extension~\cite{DBLP:conf/ijcnn/TijaniMZJD21}, 
distillation-based methods~\cite{jeong2018communication,li2019fedmd} and
Prototypical Contrastive \gls{FL}~\cite{DBLP:journals/corr/abs-2109-12273}.

Mixing local and global models has been explored
by~\cite{deng2020adaptive}, where a scalar \(\alpha\) is optimized to combine
global and local models. In~\cite{peterson2019private} the authors propose to
use \gls{MoE}~\cite{DBLP:journals/neco/JacobsJNH91} and learn a gating
function that weighs a local and global expert to enhance user privacy. This
work is developed further in~\cite{DBLP:journals/corr/abs-2010-02056}, where
the authors use a gating function with larger capacity to learn a 
personalized model when client data is
non-\gls{IID}. We differ in using cluster models as expert models, and by 
evaluating our method on datasets with different non-\gls{IID} characteristics.

Recent work has studied clustering in \gls{FL} settings for non-\gls{IID}
data~\cite{DBLP:conf/nips/GhoshCYR20,DBLP:journals/corr/abs-2012-03788,briggs2020federated}.
In~\cite{DBLP:conf/nips/GhoshCYR20} the authors implement a clustering
algorithm for handling non-\gls{IID} data in form of covariate shift. Their proposed
algorithm learns one global model per cluster with a central parameter server,
using the training loss of global models on local data of clients to perform
cluster assignment. In their work, they only perform clustering in the last layer
and aggregate the rest into a single model. If a global model cluster is unused
for some communication rounds, the global cluster model is removed from the
list to reduce communication overhead. However, this means that a client
cannot use other global cluster models to increase performance.
\section{Discussion}
We adapted the inspiring work
by~\cite{DBLP:conf/nips/GhoshCYR20} to work better in our setting and efficiently learned expert
models for non-IID client data. Sending all cluster models in each iteration introduces more
communication overhead. We addressed this by removing converged cluster models
from the set of selectable cluster models in~\Cref{alg:fl.server}, although
this is not used in our main results. This only affects the
result to a minor degree, but has a larger effect on training time due to
wasting client updates on already converged models. Another improvement is 
the reduces complexity in the cluster assignment step. A notable difference
between our work and \gls{IFCA} is that we share the \emph{all} weights, as
opposed to only the last layer in~\cite{DBLP:conf/nips/GhoshCYR20}. These
differences increase the communication overhead further, but this has
not been our priority and we leave this for future work.
\section{Conclusion}
In this paper, we have investigated personalization in a distributed and
decentralized
\gls{ML} setting where the data generated on the clients is heterogeneous with
non-\gls{IID} characteristics. We noted that neither
 \gls{FedAvg} nor state-of-the-art
solutions achieve high performance in this setting.
To address this problem, we proposed a practical framework of
\gls{MoE} using cluster models and local models as expert models and improved
 the adaptiveness of the expert models by balancing exploration and
 exploitation. Specifically, we used a \gls{MoE}~\cite{DBLP:journals/corr/abs-2010-02056} to
make better use of the cluster models available in the clients and added a
local model. We showed that~\gls{IFCA}~\cite{DBLP:conf/nips/GhoshCYR20} does not
work well in our setting, and inspired by the \gls{MAB} field, added
an \gls{eps}-greedy exploration~\cite{DBLP:conf/nips/Sutton95} method to
improve the adaptiveness of the cluster models which increased their usefulness
in the~\gls{MoE}.
We evaluated our method on three datasets representing different
 non-\gls{IID} settings, and found that our approach achieve superior
 performance in two of the datasets, and is robust in the third. Even though we
 tune our algorithm and hyperparameters in the \gls{IID} setting, it
 generalizes well in non-\gls{IID} settings or with varying number of cluster
 models --- a testament to its robustness. For example, for \cifar{} we see an
 average accuracy improvement of \SI{29.78}{\percent} compared to \gls{IFCA} and
\SI{4.38}{\percent} compared to a local model
 in the pathological non-\gls{IID} setting. Furthermore, our approach improved
 the inter-client accuracy variance with \SI{60.38647342995169}{\percent} compared
 to \gls{IFCA}, which indicates improved fairness,
 but \SI{60.97560975609756}{\percent} worse than a local model.

In real-world scenarios data is distributed and
often displays non-\gls{IID} characteristics, and we consider
personalization to be a very important direction of research.
Finding clusters of similar clients to make learning more efficient is still
an open problem. We believe there is potential to improve the convergence of
the cluster models further, and that privacy, security and system
aspects provide interesting directions for future work.
\cleardoublepage
\section*{Ethical Statement}
There are no ethical issues.
\section*{Acknowledgment}
This work was partially supported by the \gls{WASP} 
funded by the Knut and Alice Wallenberg Foundation.

The computations were enabled by resources provided
by the \gls{SNIC}, partially funded
by the Swedish Research Council through grant agreement no. 2018--05973.

We thank all reviewers who made suggestions that helped improve
and clarify this manuscript, especially Dr.~A.~Alam, F.~Cornell,
Dr.~R.~Gaigalas, T.~Kvernvik, C.~Svahn, F.~Vannella,
Dr.~H.~Shokri~Ghadikolaei, D.~Sandberg and Prof.~S.~Haridi.

\bibliographystyle{named}
\bibliography{references/main}
\balance

\end{document}